\documentclass[twocolumn,10pt]{article}

\usepackage[letterpaper, top=1in, bottom=1in, left=0.75in, right=0.75in]{geometry}
\usepackage{times}
\usepackage[hyphens]{url}
\usepackage{graphicx}
\usepackage{natbib}
\usepackage{caption}
\usepackage{hyperref}       % hyperlinks (must be loaded for cross-referencing)
\hypersetup{
    colorlinks=true,   % Removes boxes and colors the text instead
    linkcolor=blue,    % Color for internal links (equations, figures, etc.)
    filecolor=blue,    % Color for local file links
    urlcolor=blue,     % Color for linked URLs
    citecolor=blue     % Color for bibliography citations
}

\usepackage{algorithm}
\usepackage[french,english]{babel}
\usepackage{amsmath}
\usepackage{amsthm}
\usepackage{amsfonts}
\usepackage{booktabs, threeparttable}
\usepackage{multirow}
\usepackage{multicol}
\usepackage{algpseudocode}
\usepackage{xcolor,colortbl}
\usepackage{array}
\usepackage{siunitx}
\usepackage{rotating}
\usepackage{adjustbox}
\usepackage[normalem]{ulem}

\makeatletter
\def\And{\end{tabular}\hfil\linebreak[0]\hfil\begin{tabular}[t]{c}}
\def\@maketitle{%
  \newpage
  \null
  \vskip 2em%
  \begin{center}%
  \let \footnote \thanks
    {\hrule height 1pt\vskip 0.2in
     \LARGE \bfseries \@title \par
     \vskip 0.2in\hrule height 1pt}%
    \vskip 1.5em%
    {\large
      \lineskip .5em%
      \begin{tabular}[t]{c}%
        \@author
      \end{tabular}\par}%
    \vskip 1em%
  \end{center}%
  \par
  \vskip 1.5em}
\makeatother

\title{Modern Structure-Aware Simplicial Spatiotemporal Neural Network}

\author{
 Zhaobo HU \\
 SAMOVAR, T\'el\'ecom SudParis \\
 Institut Polytechnique de Paris\\
 \texttt{zhaobo.hu@telecom-sudparis.eu} \\
 \And
 Vincent Gauthier \\
 SAMOVAR, T\'el\'ecom SudParis \\
 Institut Polytechnique de Paris\\
 \texttt{vincent.gauthier@telecom-sudparis.eu} \\
 \And
 Mehdi Naima \\
 CNRS -- LIP6\\
 Sorbonne Université\\
 \texttt{mehdi.naima@lip6.com} \\
}

\begin{document}

\maketitle

\begin{abstract} 
Spatiotemporal modeling has evolved beyond simple time series analysis to become fundamental in structural time series analysis. While current research extensively employs graph neural networks (GNNs) for spatial feature extraction with notable success, these networks are limited to capturing only pairwise relationships, despite real-world networks containing richer topological relationships. Additionally, GNN-based models face computational challenges that scale with graph complexity, limiting their applicability to large networks. To address these limitations, we present Modern Structure-Aware Simplicial SpatioTemporal neural network (ModernSASST), the first approach to leverage simplicial complex structures for spatiotemporal modeling. Our method employs spatiotemporal random walks on high-dimensional simplicial complexes and integrates parallelizable Temporal Convolutional Networks to capture high-order topological structures while maintaining computational efficiency. Our source code is publicly available on GitHub\footnote{Code is available at: \url{https://github.com/ComplexNetTSP/ST_RUM}}.
\end{abstract}

\section{Introduction}

The last decade has witnessed remarkable advancements in graph neural networks (GNNs) \cite{defferrard2016convolutional, kipf2016semi, veličković2018graph}, which leverage message propagation and aggregation mechanisms to enable nodes to learn comprehensive neighborhood features, demonstrating exceptional performance in node and graph classification tasks. The integration of temporal models with GNNs has revolutionized spatiotemporal analysis \cite{li2018diffusion, wu2019graph, gao2021equivalence}, particularly in modeling complex systems like sensor networks and transportation infrastructures. However, a fundamental limitation persists: both traditional GNNs and current spatiotemporal models predominantly focus on pairwise relationships, overlooking the rich higher-dimensional interactions that characterize many real-world systems \cite{battiston2020,bick2023}, such as social network cliques, clustered sensor deployments, and aerodynamic wake effects in offshore wind farm configurations.

Simplicial complexes \cite{hajij2023topological, papillon2023architectures, schaub2020random} represent a sophisticated generalization of traditional graphs, consisting of $k$-simplices of varying orders where each order captures increasingly complex geometric structures: 0-simplices (vertices), 1-simplices (edges), 2-simplices (triangles), and 3-simplices (tetrahedrons) and so on. Each $k$-simplex maintains four distinct neighboring relationships: boundary connections to $k-1$-simplices, co-boundary links to $k+1$-simplices, and both upper and lower adjacencies to other $k$-simplices, creating a rich topological structure. Recent Simplicial Neural Networks (SNNs) \cite{bodnar2021weisfeiler, bunch2020simplicial, gurugubelli2024sann} extend GNN paradigms to these higher-dimensional structures, leveraging all four neighboring relationships to capture substantially richer topological information. However, this enhanced capability comes with significant computational overhead through multiple learnable parameters per layer and large neighborhood matrices, substantially increasing training time and computational requirements. While traditional graph-based spatiotemporal models operate with complexity $\mathcal{O}(TE)$ where $T$ represents time steps and $E$ represents edge count, implementing SNNs would result in $\mathcal{O}(T(E + F))$ where $F$ represents triangular faces, making practical implementation considerably more resource-intensive.

To address these computational challenges while preserving rich topological information, we propose the Modern Structure-Aware Simplicial Spatiotemporal neural network (ModernSASST), which abandons performance-degrading recurrent neural networks and message-passing mechanisms in favor of parallelizable Temporal Convolutional Networks (TCN) \cite{luo2024moderntcn, bai2018empirical} and spatiotemporal random walks. This approach leverages the inherent parallelism of convolutional operations while utilizing random walk-based exploration to effectively capture complex topological dependencies within simplicial complexes, achieving superior computational efficiency compared to traditional recurrent architectures while maintaining the ability to model complex spatiotemporal dependencies.

\subsection*{Contributions}
The main contributions are summarized as follows:
\begin{itemize}
\item[$\bullet$] To the best of our knowledge, we are the first to leverage high-dimensional simplicial complex structures for spatiotemporal modeling. We propose ModernSASST, a simple yet practical approach for incorporating topological awareness into spatiotemporal analysis.

\item[$\bullet$] We conduct comprehensive experiments on three diverse real-world datasets spanning energy, environmental, and transportation domains, demonstrating ModernSASST's effectiveness across varied spatiotemporal characteristics.

\item[$\bullet$] Although primarily designed for prediction, ModernSASST also achieves competitive performance in spatiotemporal data imputation, showcasing the versatility of our topological modeling approach.
\end{itemize}

\section{Related Works}
Random walk-based graph embedding methods have established fundamental approaches for learning node representations. DeepWalk \cite{perozzi2014deepwalk} pioneered the field by treating truncated random walks as sentences for Skip-gram models, while node2vec \cite{grover2016node2vec} introduced biased random walks with parameters controlling breadth-first and depth-first search strategies. LINE \cite{tang2015line} extended these concepts to large-scale networks by preserving first and second-order proximities. The extension to higher-order structures represents a cutting-edge frontier, with Schaub et al. \cite{schaub2020random} providing foundational work on random walks on simplicial complexes and the normalized Hodge 1-Laplacian, establishing connections between Hodge Laplacians and random walk dynamics. Building on these foundations, Billings et al. \cite{billings2019simplex2vec} introduced Simplex2Vec, which performs random walks on Hasse diagrams of simplicial complexes to preserve higher-order topological information, while Hacker \cite{hacker2020k} developed k-simplex2vec as a direct extension of node2vec to higher-dimensional simplices. Topological deep learning architectures like Simplicial Neural Networks \cite{ebli2020simplicial} and Cell Complex Neural Networks \cite{bodnar2021weisfeiler} have extended GNNs to simplicial complexes, enabling message passing that respects topological structure and higher-order interactions beyond pairwise relationships.

\section{Preliminary}

\subsection{Problem Statement}
\label{sec:Problem Statement}
Consider a network with $N$ nodes, where each node $i$ generates an $f$-dimensional feature vector $x_t^i \in \mathbb{R}^f$ at time step $t$. The network state at time $t$ is represented by $X_t \in \mathbb{R}^{N \times f}$, while a sequence spanning time interval $T$ is denoted as $X_{t:t+T}$. The network topology is characterized by adjacency matrix $A \in \mathbb{R}^{N \times N}$, where $a_{ij}$ indicates connection strength between nodes $i$ and $j$. Our work addresses two fundamental spatiotemporal challenges:
\begin{enumerate}
\item \textbf{Spatiotemporal Prediction}: Given historical observations $X_{t-W:t} \in \mathbb{R}^{N \times f \times W}$ spanning $W$ time steps, predict the network's future evolution $X_{t+1:t+H} \in \mathbb{R}^{N \times f \times H}$ over the next $H$ time steps.
\begin{equation*}
   Y_{t+1:t+H} = F_{\theta}(X_{t-W:t}, A)
\end{equation*}
\item \textbf{Spatiotemporal Imputation}: Reconstruct missing values within a partial observation window $X_{t:t+W} \in \mathbb{R}^{N \times f \times W}$, where certain elements are unobserved due to sensor failures or sampling limitations. Let $M \in \{0,1\}^{N \times f \times W}$ be a binary mask matrix where $M_{i,j,k} = 1$ indicates missing values.
\begin{equation*}
   \hat{X}_{t:t+W} = F_{\theta}(X_{t:t+W}, A, M)
\end{equation*}
\end{enumerate}
Both tasks require modeling complex dependencies across spatial and temporal dimensions through a parameterized function $F_\theta(\cdot)$ that integrates graph structural information with temporal dynamics, where $\theta$ represents the learnable parameters.
 
\subsection{Simplicial Complex}
Let $V = \{v_0, v_1, \ldots, v_N\}$ be a non-empty vertex set of cardinality $N + 1$. A simplicial complex $\mathcal{K}$ is a collection of non-empty subsets of $V$ that satisfies the closure property: if $\sigma \in \mathcal{K}$ and $\tau \subset \sigma$, then $\tau \in \mathcal{K}$. An element $\sigma = \{v_0, v_1, \ldots, v_k\} \in \mathcal{K}$ with cardinality $k + 1$ is called a $k$-simplex (simplex of order $k$), where 0-simplices represent vertices, 1-simplices edges, 2-simplices triangles, and so on. We denote by $|\sigma|$ the order of $\sigma$ (i.e the size of $\sigma$ minus one) and by $N_k$ the number of $k$-simplices in $\mathcal{K}$. A simplicial complex enables the representation of adjacency relationships among simplices in the following way:

\begin{figure*}[t]
    \centering
    \includegraphics[width=\textwidth,keepaspectratio]{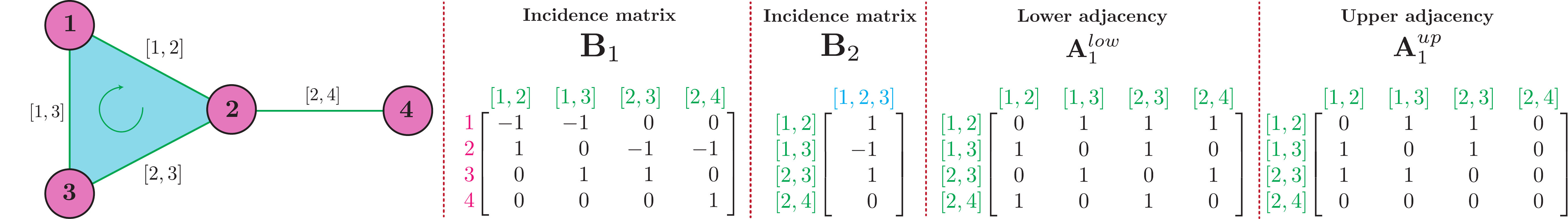}
    \caption{On the left we depicted the complex simplicial $\mathcal{K}$ on $4$ vertices, where $\mathbf{B}_1$ and $\mathbf{B}_2$ are boundary matrices of the complex simplicial $\mathcal{K}$, then lower and upper adjacency matrices of the complex simplicial $\mathcal{K}$}
    \label{fig:SC}
\end{figure*}

\begin{enumerate}
   \item boundary adjacencies $\mathcal{B}(\sigma) = \{\tau \mid \tau \prec \sigma\}$ containing all $(k-1)$-dimensional faces, e.g., for triangle $\{1,2,3\}$: $\mathcal{B}(\sigma) = \{ \{1,2\}, \{2,3\}, \{1, 3\} \}$.
   \item co-boundary adjacencies $\mathcal{C}(\sigma) = \{\tau \mid \sigma \prec \tau\}$ comprising all $(k+1)$-simplices containing $\sigma$, e.g., for edge $\{1,2\}$: $\mathcal{C}(\sigma) = \{ \{1,2,3\}\}$.
   \item lower adjacencies $\mathcal{N}_{\downarrow}(\sigma) = \{\tau \mid \exists \delta, \delta \prec \tau \wedge \delta \prec \sigma\}$ connecting simplices sharing common faces, e.g., $\mathcal{N}_{\downarrow}(\{1,2\}) = \{ \{1,3\}, \{2,3\}, \{2,4\}\}$.
   \item upper adjacencies $\mathcal{N}_{\uparrow}(\sigma) = \{\tau \mid \exists \delta, \tau \prec \delta \wedge \sigma \prec \delta\}$ linking simplices contained in common higher-dimensional simplices, e.g., $\mathcal{N}_{\uparrow}(\{1,2\}) = \{ \{1,3\}, \{2,3\}\}$.
\end{enumerate}

These boundary relations of $k$-simplices  are encoded through the oriented incidence matrix $\mathbf{B}_k \in \mathbb{R}^{N_{k-1} \times N_k}$ where an additional reference orientation on the simplices is required see~\cite{schaub2020random}. Lower and upper adjacencies of $k$-simplices can be encoded through matrices $ \mathbf{A}^{up}_k, \mathbf{A}^{low}_k \in \mathbb{R}^{N_{k} \times N_k}$ where $(\mathbf{A}^{up}_k)_{ij} = 1$ if simplices $s_i$ and $s_j$ are upper adjacent  and $0$ otherwise. The lower adjacency is defined similarly. Finally, we define the adjacency matrix $\mathbf{A}_k$ of order $k$ as:
\begin{equation*}
(\mathbf{A}_k)_{ij} = 
    \begin{cases}
        1 & \text{ if $(\mathbf{A}^{up}_k)_{ij} = 1$ or $(\mathbf{A}^{low}_k)_{ij} = 1$}\\
        0 & \text{ otherwise.}
    \end{cases}
\end{equation*}

An example of a simplical complex together with its adjacency relationships is given on Figure~\ref{fig:SC}.

The $k$-th order Hodge Laplacian \cite{schaub2020random}, which generalizes the standard graph Laplacian to higher-order structures, is defined as:
\begin{equation*}
\mathbf{L}_k  = \mathbf{B}_k^T \mathbf{B}_k + \mathbf{B}_{k+1}\mathbf{B}_{k+1}^T.
\end{equation*}
The Hodge Laplacian combines both lower and upper components. In this study, we restrict our analysis to 2-order simplicial complexes, comprising 0-simplices (vertices), 1-simplices (edges), and 2-simplices (triangles). The 2-simplices are constructed through identification of 3-cliques within the graph structure, where a 3-clique is a complete subgraph of three vertices $\{v_i, v_j, v_k\} \subset V$ with all pairwise connections $\{(v_i,v_j), (v_j,v_k), (v_i,v_k)\} \subset E$ as it has been done in \cite{bodnar2021weisfeiler}. 

Simplicial Neural Networks (SNNs) operate on simplicial complexes by propagating information across these adjacency structures. However, they suffer from performance degradation due to the substantial number of learnable parameters required for multiple adjacency matrices. To address these limitations, we propose a random-walk based approach that leverages the structural properties of simplicial complexes without requiring extensive parameter learning.

\section{Model design}
\subsection{Random walk as Spatial Encoder}
Random walk-based graph embedding methods have been extensively studied \cite{perozzi2014deepwalk, grover2016node2vec, dong2017metapath2vec, he2019hetespaceywalk, tang2015line}. Notable works include DeepWalk \cite{perozzi2014deepwalk} and Node2Vec \cite{grover2016node2vec}, with recent developments exploring random walks on higher-order structures and simplicial complexes \cite{billings2019simplex2vec, hacker2020k}. We define the full adjacency matrix as $\mathcal{A} \in \mathbb{R}^{(N+E+T) \times (N+E+T)}$, where $N$, $E$, and $T$ represent the number of nodes, edges, and triangles respectively. Using the adjacency and boundary matrices from the previous section, we construct two full adjacency matrix variants:

\begin{equation*}
\hat{\mathcal{A}}_1 = \begin{bmatrix}
\mathbf{A}_0 & \mathbf{B}_1 & 0 \\
\mathbf{B}_1^T & \mathbf{A}_1 & \mathbf{B}_2 \\
0 & \mathbf{B}_2^T & \mathbf{A}_2
\end{bmatrix}, \\
\hat{\mathcal{A}}_2 = \begin{bmatrix}
0 & \mathbf{B}_1 & 0 \\
\mathbf{B}_1^T & 0 & \mathbf{B}_2 \\
0 & \mathbf{B}_2^T & 0
\end{bmatrix}
\end{equation*}

Since the Hodge Laplacians $\mathbf{L}_k$ cannot be directly used as transition matrices for random walk sampling. Therefore, for practical implementation, we replace these Laplacians with their corresponding adjacency matrices $\mathbf{A}_k$. Additionally, we remove the negative signs from $\mathbf{B}_1$ and $\mathbf{B}_2$ to ensure non-negative transition probabilities. It is worth noting that the orientation in simplicial complexes differs from the direction in graphs, while graph direction indicates allowed transition paths, simplicial orientation is a mathematical construct for consistent ordering of vertices \cite{schaub2020random}. 

\begin{figure*}[htbp]
  \centering
  \includegraphics[width=\textwidth,keepaspectratio]{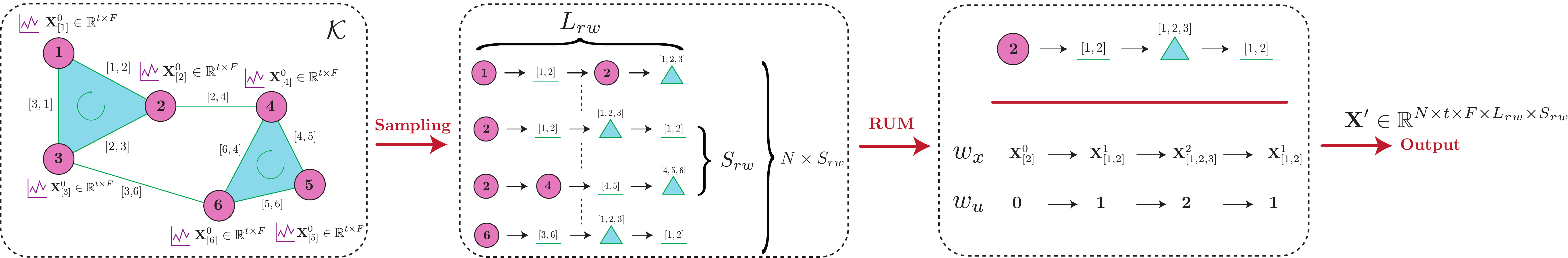}
  \caption{RUM}
  \label{fig:rum_sampling}
\end{figure*}

Matrix $\mathcal{\hat{A}}_1$ enables same-order transitions via diagonal terms, while $\mathcal{\hat{A}}_2$ constrains transitions to different orders only. When edges or triangles dominate, $\mathcal{\hat{A}}_1$ may trap walkers in higher-dimensional simplices, while $\mathcal{\hat{A}}_2$ provides better mixing across dimensional levels. We employ Random Walk with Unifying Memory (RUM) \cite{wang2025non}, which processes semantic and topological trajectories through GRU networks \cite{chung2014empirical}.\\

\noindent\textbf{Random walks on simplicial complexes}\quad An unbiased random walk $w$ on simplicial complex $\mathcal{K}$ is defined as a sequence of simplices $w = (s_0, s_1, \ldots)$ with transition probability:
\begin{equation*}
P(s_j \mid s_i) = \begin{cases}
        1/D(s_i) & \text{if } s_j \in \mathcal{N}(s_i)\\
        0 & \text{ otherwise.}
    \end{cases}
\end{equation*}

\noindent where $D(s_i) = \sum_k \mathcal{\hat{A}}_{ik}$ represents the degree of simplex $s_i$. For biased random walks, we introduce order-aware transition probabilities:
\begin{equation*}
P(s_j \mid s_i) = \begin{cases}
        \alpha_{|s_j|}/ \sum_{s_k \in \mathcal{N}(s_i)} \alpha_{|s_k|} & \text{if } s_j \in \mathcal{N}(s_i)\\
        0 & \text{ otherwise.}
    \end{cases}
\end{equation*}

\noindent where $\alpha_{|s|} = 1/N_{|s|}$ penalizes transitions to higher-count orders, preventing walkers from being trapped in dominant structures. If we use $\hat{\mathcal{A}}_1$ as adjacency matrix, the neighborhood of simplex $s_i$ is define as $\mathcal{N}(s_i) = \mathcal{N}_{\uparrow}(s_i) \cup \mathcal{N}_{\downarrow}(s_i) \cup \mathcal{B}(s_i) \cup \mathcal{C}(s_i)$, and $\mathcal{N}(s_i) =  \mathcal{B}(s_i) \cup \mathcal{C}(s_i)$ if we use $\hat{\mathcal{A}}_2$ as adjacency matrix.

We record the semantics trajectory as $\omega_x(w) = (\mathbf{X}^{k}_0, \mathbf{X}^{k}_1, \ldots, \mathbf{X}^{k}_l)$, where $\mathbf{X}^{k}_i$ represents the feature vector of the $k$-dimensional simplex visited at step $i$, and focus on finite-length $l$-step random walks sampled using Deep Graph Library \cite{wang2019deep}.\\

\noindent\textbf{Anonymous experiment}\quad We employ \textit{anonymous experiment} \cite{micali2016reconstructing}, $\omega_u(w) : \mathbb{R}^l \to \mathbb{R}^l$ that records \textbf{the first unique occurrence of a simplex in a walk}, labeling simplices with the number of unique simplices traversed (see Figure~\ref{fig:rum_sampling}).

\subsection{ModernSASST}
To address the computational overhead and limitations of the dual-GRU structure in the original RUM framework, we recognize that RNN-based networks lack parallel computation capabilities. Therefore, we reconstruct ModernTCN \cite{luo2024moderntcn} to provide a variant specifically tailored for spatiotemporal graph networks, termed ModernSASST. Benefiting from the channel-independent approach in time series forecasting \cite{luo2024moderntcn, Yuqietal-2023-PatchTST}, we circumvent the dual-GRU structure entirely. This paradigm allows each trajectory component to be processed independently and in parallel, eliminating separate semantic and topological GRU encoders while maintaining representational capacity for capturing complex spatiotemporal patterns.\\

\begin{figure}[htbp]
    \centering
    \includegraphics[width=\columnwidth,keepaspectratio]{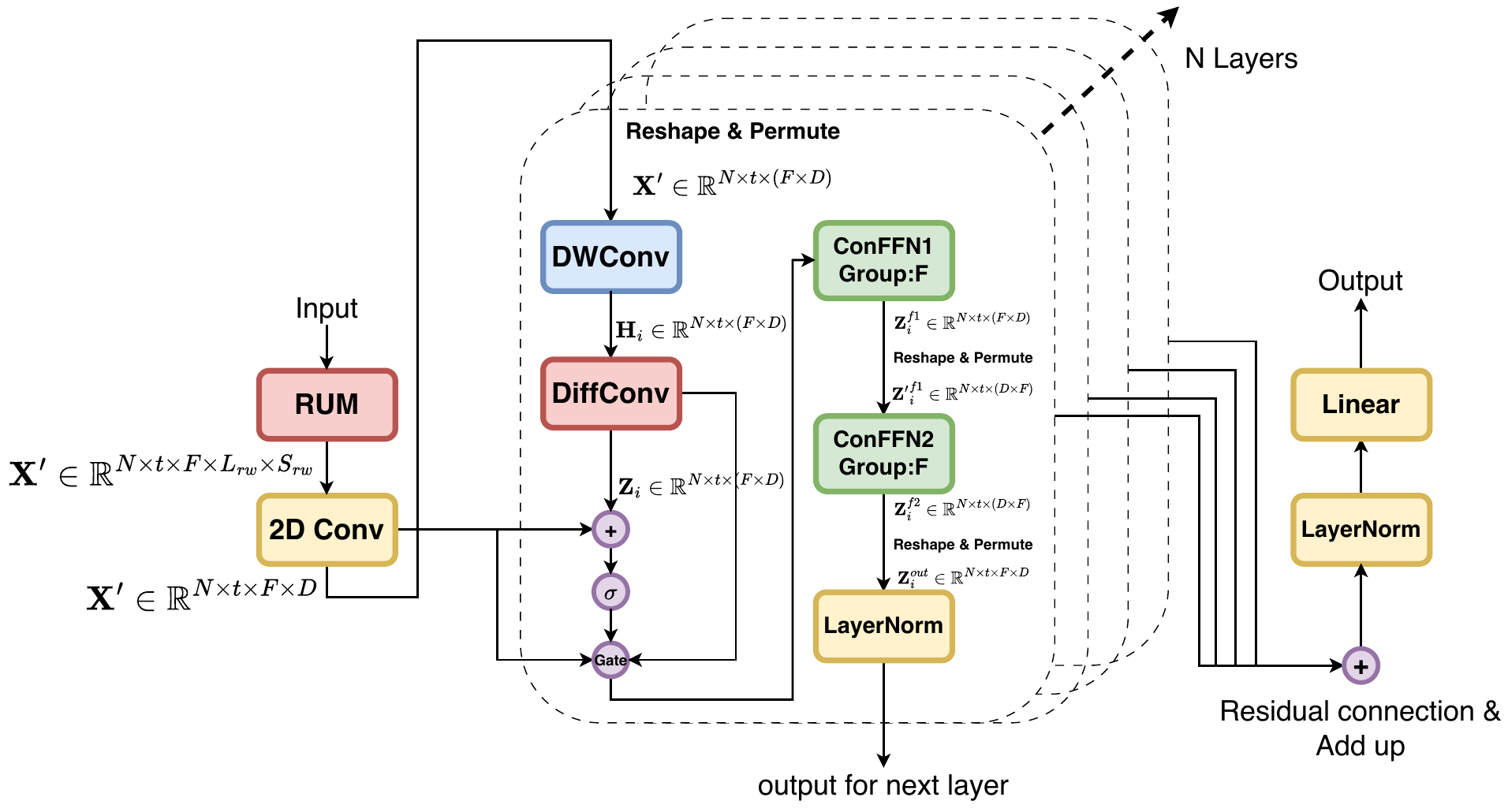}
    \caption{Model}
    \label{fig:model}
\end{figure}

\noindent\textbf{Random walk features extraction}\quad We consider the input as $\mathbf{X} \in \mathbb{R}^{N \times t \times F}$, where $N$, $t$, and $F$ represent nodes, temporal steps and feature dimensions respectively. We also incorporate edge features $\mathbf{X}^1 \in \mathbb{R}^{E \times F_1}$ and triangle features $\mathbf{X}^2 \in \mathbb{R}^{T \times F_2}$. The expansion operation adjusts edge and triangle features to match the temporal steps and node feature dimensions, followed by concatenation along the first dimension (simplices count):
\begin{equation*}
\mathbf{X} = \text{Concat}(\mathbf{X}, \text{Expand}(\mathbf{X}^1), \text{Expand}(\mathbf{X}^2))
\end{equation*}
The RUM framework performs random walk sampling on the unified features for each node at every time step, with parameters walk length $L_{rw}$ and number of samples $S_{rw}$. Our input $\mathbf{X}$ is transformed to $\mathbf{X}' \in \mathbb{R}^{N \times t \times F \times L_{rw} \times S_{rw}}$. This substantial dimensionality increase poses computational challenges, potentially causing GPU memory overflow or degraded training efficiency.
We perform an unsqueezing operation, transforming to $\mathbf{X}' \in \mathbb{R}^{N \times t \times F \times 1 \times L_{rw} \times S_{rw}}$. We employ a 2D convolution as the STEM layer with kernel size $[L^{'}, S_{rw}]$. Following Flatten and Convolution transformations, $\mathbf{X}'$ is reshaped to $\mathbf{X}' \in \mathbb{R}^{N \times t \times F \times D}$, where $D$ represents the output embedding dimension. This design compresses the random walk sampling dimensions $(L_{rw} \times S_{rw})$ into unified feature representation through the 2D convolutional STEM layer, reducing computational complexity while preserving topological information. Details are given in Figure~\ref{fig:model}.\\

\noindent\textbf{DWConv}\quad The DWConv employs specific kernel sizes to learn temporal features. Consistent with the currently popular channel-independent methodology, we modify the DWConv to operate at the node-wise with both channel and feature independence, enabling each node's channels and features to be processed independently. This architectural adaptation aligns with the channel-independent paradigm prevalent in modern time series forecasting, where individual channels are treated as separate entities without cross-channel interactions.\\

\noindent\textbf{Gate fusion}\quad We only extract spatial features at the STEM layer, as continuing RUM-based extraction in deeper layers would incur substantial computational overhead. To address this, we adopt the self-adaptive learnable adjacency matrix from Graph Wavenet \cite{wu2019graph} by randomly initializing two node embedding dictionaries $\mathbf{U}, \mathbf{V} \in \mathbb{R}^{N \times c}$. The self-adaptive adjacency matrix is:
\begin{equation*}
\tilde{\mathbf{A}}_{adp} = \text{SoftMax}(\sigma(\mathbf{U} \mathbf{V}^T))
\end{equation*}
\noindent where $\mathbf{U}$ and $\mathbf{V}$ represent source and target node embeddings. The activation function $\sigma(\cdot)$ eliminates weak connections and SoftMax normalizes the matrix, serving as a transition matrix for latent diffusion without repeated random walk sampling overhead. Following Graph WaveNet and DCRNN, the diffusion process is formulated as:
\begin{equation*}
    Z_i = \sum_{k=0}^{K} \tilde{\mathbf{A}}_{adp}^k H_i \mathbf{W}_k
\end{equation*}
where $Z_i$ denotes the diffusion output, $\tilde{\mathbf{A}}_{adp}^k$ captures $k$-hop neighborhood information, $H_i$ is the DWConv output, $\mathbf{W}_k$ is the learnable weight matrix, and $K=2$ controls the spatial receptive field. To fuse low-level and high-level spatial representations, we employ a gating mechanism:
\begin{equation*}
    \text{gate} = Sigmoid(\mathbf{X}' + Z_i)
\end{equation*}

\begin{equation*}
    Z'_i = \text{gate} \odot \mathbf{X}' + (1 - \text{gate}) \odot Z_i
\end{equation*}
\noindent where $\mathbf{X}'$ is the STEM layer output, $\odot$ indicates element-wise multiplication, and $Z'_i$ adaptively balances spatial information. \\

\noindent\textbf{ConvFFN}\quad Since DWConv operates with channel and feature independence in our spatiotemporal framework, ConvFFN must facilitate information mixing across channel and feature dimensions. A unified ConvFFN approach leads to higher computational complexity and suboptimal performance. Therefore, we decouple the unified ConvFFN into ConvFFN1 and ConvFFN2 by replacing pointwise convolutions with grouped pointwise convolutions using different group parameters. ConvFFN1 learns enhanced feature representations for individual channels, focusing on intra-channel feature interactions, while ConvFFN2 captures cross-channel dependencies within each feature dimension, enabling inter-channel communication. For detailed descriptions of DWConv and ConvFFN, we refer readers to the ModernTCN paper \cite{luo2024moderntcn}\\

\noindent\textbf{Necessity of graph-wise layernorm}\quad We employ graph-wise layernorm \cite{ba2016layernormalization} to normalize residual connection outputs throughout the architecture. Random walk-based methods function analogously to data augmentation techniques, enhancing generalization but potentially introducing training instability. We adopt graph-level rather than channel-wise normalization because certain nodes may exhibit severe distributional bias following random walk sampling, which could compromise training stability. This strategy maintains consistent learning dynamics when dealing with variability from the random walk sampling process.\\

\section{Experiments}
\begin{table*}[t]
\centering
\begin{tabular}{l|cc|cc|cc}
\toprule
& \multicolumn{2}{c|}{SDWPF} & \multicolumn{2}{c|}{AQI} & \multicolumn{2}{c}{METR-LA} \\
\cmidrule{2-3} \cmidrule{4-5} \cmidrule{6-7}
& MAE & RMSE & MAE & RMSE & MAE & RMSE \\
\midrule
VAR & 95.97 $\pm$ 3.76 & 159.25 $\pm$ 0.37 & 31.61 $\pm$ 0.08 & 47.41 $\pm$ 0.34 & 4.17 $\pm$ 0.03 & 7.02 $\pm$ 0.04\\
FC-LSTM & 87.55 $\pm$ 1.71 & 161.09 $\pm$ 3.86 & 25.91 $\pm$ 0.22 & 41.10 $\pm$ 0.30 & 3.55 $\pm$ 0.01& 6.92 $\pm$ 0.01\\
ModernTCN & 90.77 $\pm$ 0.79 & 168.17 $\pm$ 0.35 & 22.32 $\pm$ 0.02& 38.76 $\pm$ 0.19& 3.64 $\pm$ 0.01& 7.34 $\pm$ 0.03\\
\midrule
DCRNN & 93.67 $\pm$ 1.60 & 166.48 $\pm$ 2.68 & 22.57 $\pm$ 0.05 & 39.23 $\pm$ 0.02 & 3.23 $\pm$ 0.01 & 6.27 $\pm$ 0.01 \\
GWavenet & 84.85 $\pm$ 2.68 & 152.49 $\pm$ 0.74 & 21.15 $\pm$ 0.10 & 37.07 $\pm$ 0.34 & 3.17 $\pm$ 0.03 & 6.26 $\pm$ 0.04 \\
SGP & 89.64 $\pm$ 0.43 & 163.86 $\pm$ 1.10 & 22.10 $\pm$ 0.23 & 38.98 $\pm$ 0.60 & 3.14 $\pm$ 0.01 & 6.31 $\pm$ 0.03\\
\midrule
\textbf{Ours - Undirected Graph} & 81.74 $\pm$ 0.81 & 152.10 $\pm$ 1.54 & \textcolor{blue}{20.89 $\pm$ 0.22} & 36.07 $\pm$ 0.12 & 3.01 $\pm$ 0.02 & \textcolor{blue}{6.02 $\pm$ 0.04} \\
\textbf{Ours - Directed Graph} & - & - & - & - & \textcolor{blue}{3.01 $\pm$ 0.01} & 6.04 $\pm$ 0.04 \\
\textbf{Ours - $1-order$ structure} & \textcolor{blue}{80.75 $\pm$ 0.70} & \textcolor{blue}{150.36 $\pm$ 0.74} & 20.94 $\pm$ 0.27 & \textcolor{blue}{35.98 $\pm$ 0.41} & 3.01 $\pm$ 0.04 & 6.00 $\pm$ 0.02 \\
\textbf{Ours - $2-order$ structure} & \textcolor{red!80!black}{\textbf{79.63 $\pm$ 0.78}} & \textcolor{red!80!black}{\textbf{148.97 $\pm$ 1.11}} & \textcolor{red!80!black}{\textbf{20.89 $\pm$ 0.20}} & \textcolor{red!80!black}{\textbf{36.01 $\pm$ 0.26}} & \textcolor{red!80!black}{\textbf{3.00 $\pm$ 0.01}} & \textcolor{red!80!black}{\textbf{6.00 $\pm$ 0.01}}\\
\bottomrule
\end{tabular}
\caption{Forecasting performance comparison on SDWPF, AQI, and METR-LA datasets}
\label{tab:forecasting_results}
\end{table*}

\begin{table*}[t]
\centering
\begin{adjustbox}{width=\textwidth,center}
\begin{tabular}{l|cc|cc|cc}
\toprule
& \multicolumn{2}{c|}{SDWPF} & \multicolumn{2}{c|}{AQI} & \multicolumn{2}{c}{METR-LA} \\
\cmidrule{2-3} \cmidrule{4-5} \cmidrule{6-7}
& MAE & MRE (\%) & MAE & MRE (\%) & MAE & MRE (\%) \\
\midrule
BRITS & 105.90 $\pm$ 0.54 & 44.55 $\pm$ 0.23 & 28.80 $\pm$ 0.04 & 44.31 $\pm$ 0.07 & 4.65 $\pm$ 0.01 & 8.05 $\pm$ 0.01 \\
GRIN & 33.62 $\pm$ 0.46 & 14.37 $\pm$ 0.20 & \textcolor{red!80!black}{\textbf{21.51 $\pm$ 1.46}} & \textcolor{red!80!black}{\textbf{33.09 $\pm$ 2.24}} & \textcolor{red!80!black}{\textbf{2.22 $\pm$ 0.02}} & \textcolor{red!80!black}{\textbf{3.84 $\pm$ 0.04}}\\
SPIN & 83.68 $\pm$ 5.80 & 35.76 $\pm$ 2.48 & 29.04 $\pm$ 2.18 & 43.64 $\pm$ 3.92 & 5.46 $\pm$ 1.13 & 9.40 $\pm$ 1.95\\
\midrule
\textbf{Ours - Undirected Graph} & \textcolor{blue}{30.26 $\pm$ 0.83} & \textcolor{blue}{12.91 $\pm$ 0.36} & 27.66 $\pm$ 2.18 & 42.55 $\pm$ 3.35 & 2.40 $\pm$ 0.05 & 4.15 $\pm$ 0.08 \\
\textbf{Ours - Directed Graph} & - & - & - & - & \textcolor{blue}{2.37 $\pm$ 0.04} & \textcolor{blue}{4.11 $\pm$ 0.08}\\
\textbf{Ours - $1-order$ structure} & \textcolor{red!80!black}{\textbf{30.06 $\pm$ 0.87}} & \textcolor{red!80!black}{\textbf{12.83 $\pm$ 0.37}} & 27.85 $\pm$ 1.84 & 42.84 $\pm$ 2.83 & 2.38 $\pm$ 0.03 & 4.13 $\pm$ 0.05  \\
\textbf{Ours - $2-order$ structure} & 30.49 $\pm$ 1.74 & 13.01 $\pm$ 0.75 & \textcolor{blue}{27.25 $\pm$ 1.08} & \textcolor{blue}{41.89 $\pm$ 1.67} & 2.43 $\pm$ 0.05 & 4.21 $\pm$ 0.09\\
\bottomrule
\end{tabular}
\end{adjustbox}
\caption{Imputation performance comparison on SDWPF, AQI, and METR-LA datasets}
\label{tab:imputation_results}
\end{table*}

\subsection{Datasets}
Our evaluation encompasses three diverse spatiotemporal datasets with distinct graph construction methodologies. The Spatial Dynamic Wind Power Forecasting \textbf{SDWPF} dataset \cite{zhou2024sdwpf} provides wind power data from 134 turbines over 24 months with 19 dynamic features from SCADA systems and ERA5 meteorological data. The \textbf{METR-LA} dataset captures traffic patterns from 207 highway loop detectors in Los Angeles County over 4 months, while the \textbf{Air Quality} dataset records PM2.5 measurements from 437 monitoring stations across Chinese cities. For graph construction, SDWPF utilizes Delaunay triangulation for simplicial complex generation, while METR-LA and Air Quality datasets employ Gaussian kernel thresholding based on geographic distances \cite{wu2019graph, li2018diffusion}, with 3-cliques used to identify simplices. Node features represent sensor measurements, edge features encode pairwise distances, and triangle features capture geometric properties through triangle areas. Each dataset is split chronologically into $70\%$ training, $10\%$ validation, and $20\%$ testing sets.

\begin{table}[h!]
\centering
\begin{adjustbox}{width=\columnwidth,center}
\begin{tabular}{lccccl}
\toprule
Dataset & Nodes & Edges & Triangles & Sampling Frequency & Data Type \\
\midrule
SDWPF & 134 & 568  & 254 & 10 minutes & Energy \\
METR-LA & 207 & 1515  & 3293 & 5 minutes & Traffic \\
AQI & 437 & 5398  & 13680 & 1 hour & Environment \\
\bottomrule
\end{tabular}
\end{adjustbox}
\caption{Datasets description}
\label{tab:datasets}
\end{table}

\subsection{Baselines}
\noindent\textbf{Forecasting task}\quad \textbf{VAR}\cite{hamilton2020time} Vector Auto-Regression; \textbf{FC-LSTM}\cite{sutskever2014sequence} Recurrent Neural Network with fully connected LSTM hidden units; \textbf{ModernTCN}\cite{luo2024moderntcn} Modernized TCN architecture with large kernel convolutions and decoupled design for larger receptive fields; \textbf{DCRNN} \cite{li2018diffusion} Models traffic as diffusion process on directed graphs, combining graph diffusion convolutions with RNNs; \textbf{Graph WaveNet} \cite{wu2019graph} Combines graph convolutions with dilated causal convolutions and adaptive adjacency matrix for hidden spatial dependencies; \textbf{SGP}\cite{cini2023scalable} Uses randomized Echo State Networks for temporal encoding and graph shift operators for spatial propagation with constant-time scalability.\\
\noindent\textbf{Imputation task}\quad \textbf{BRITS}\cite{cao2018brits} Bidirectional RNN approach that handles missing values through bidirectional recurrent dynamics using past and future information; \textbf{GRIN}\cite{cini2022filling} Graph neural networks for multivariate time series imputation by capturing temporal dynamics and inter-variable relationships; \textbf{SPIN}\cite{marisca2022learning} Graph neural network for spatiotemporal imputation with sparse observations by capturing spatial dependencies and temporal patterns.

\subsection{Experimental setup}
We evaluate our approach on two fundamental spatiotemporal tasks: forecasting and imputation. Both tasks use consistent 12-step time windows, with forecasting predicting subsequent 12 steps and imputation focusing on out-of-sample settings. For random walk sampling, we employ the full adjacency matrix $\mathcal{\hat{A}}_1$ across all datasets, with unbiased random walks for SDWPF and biased random walks for METR-LA and AQI datasets to handle their imbalanced topological structures. Specifically, for METR-LA and SDWPF datasets, we randomly mask $5\%$ of available data and simulate sensor failures lasting $S \sim U(12, 48)$ steps with a $0.15\%$ probability for imputation task. Hyperparameter optimization uses Optuna \cite{akiba2019optuna} with random walk parameters searched within $[2, 8]$, hidden size from $[16, 32, 64]$, and dropout set to $0.1$. Learning rates are $1e^{-2}$ for SDWPF and METR-LA, $1e^{-3}$ for AQI, with StepLR scheduler. The Conv2D layer uses kernel size $[\text{2, random walk samples}]$, followed by three DWConv1D layers with kernels $[7, 5, 3]$. We train for 200 epochs with early stopping, using MAE and MSE for forecasting, MAE and MRE for imputation. All experiments run over 5 independent runs on NVIDIA L40s GPU.

\subsection{Experiment results}
Based on Tables~\ref{tab:forecasting_results} and~\ref{tab:imputation_results}, our ModernSASST framework demonstrates distinct performance patterns driven by dataset construction methodologies. The most significant improvements occur on SDWPF, where our 2-order structure consistently outperforms baselines, attributed to Delaunay triangulation creating balanced topological distribution (134 nodes, 568 edges, 254 triangles) that accurately reflects spatial relationships in wind farm configurations. This enables capturing meaningful higher-order interactions such as aerodynamic wake effects between multiple turbines. Conversely, AQI and METR-LA employ 3-clique identification from distance-based graphs, resulting in severely imbalanced topologies where triangles vastly outnumber nodes (AQI: 13,680 vs. 437; METR-LA: 3,293 vs. 207). These disproportionate structures introduce noise as many triangular relationships represent algorithmic artifacts rather than meaningful interactions, limiting performance gains despite competitive results. Task-specific analysis reveals forecasting benefits substantially from global patterns captured through higher-order topological structures for modeling cascading effects, while imputation relies more on local neighborhood information where traditional graph methods already provide substantial capacity. The ablation study in Table~\ref{tab:ablation_results} exposes contrasting sensitivity patterns. The anonymous experiment component shows moderate importance (2.73 MAE degradation in forecasting, 2.86 in imputation), indicating topological labeling provides valuable structural information. For forecasting, our model shows minimal sensitivity to self-learned adjacency and gating components since random walk sampling already incorporates sufficient spatiotemporal information, but exhibits extreme sensitivity to graph-wise LayerNorm (9.58 MAE degradation) due to distributional shifts from traversing heterogeneous simplicial structures. Imputation demonstrates opposite patterns with higher sensitivity to adaptive adjacency components (16.05 MAE degradation) as local information exchange facilitates value reconstruction, while showing reduced LayerNorm dependency because masking strategies naturally focus attention on specific missing locations.

\begin{figure*}[htbp]
    \centering
    \includegraphics[width=\textwidth,keepaspectratio]{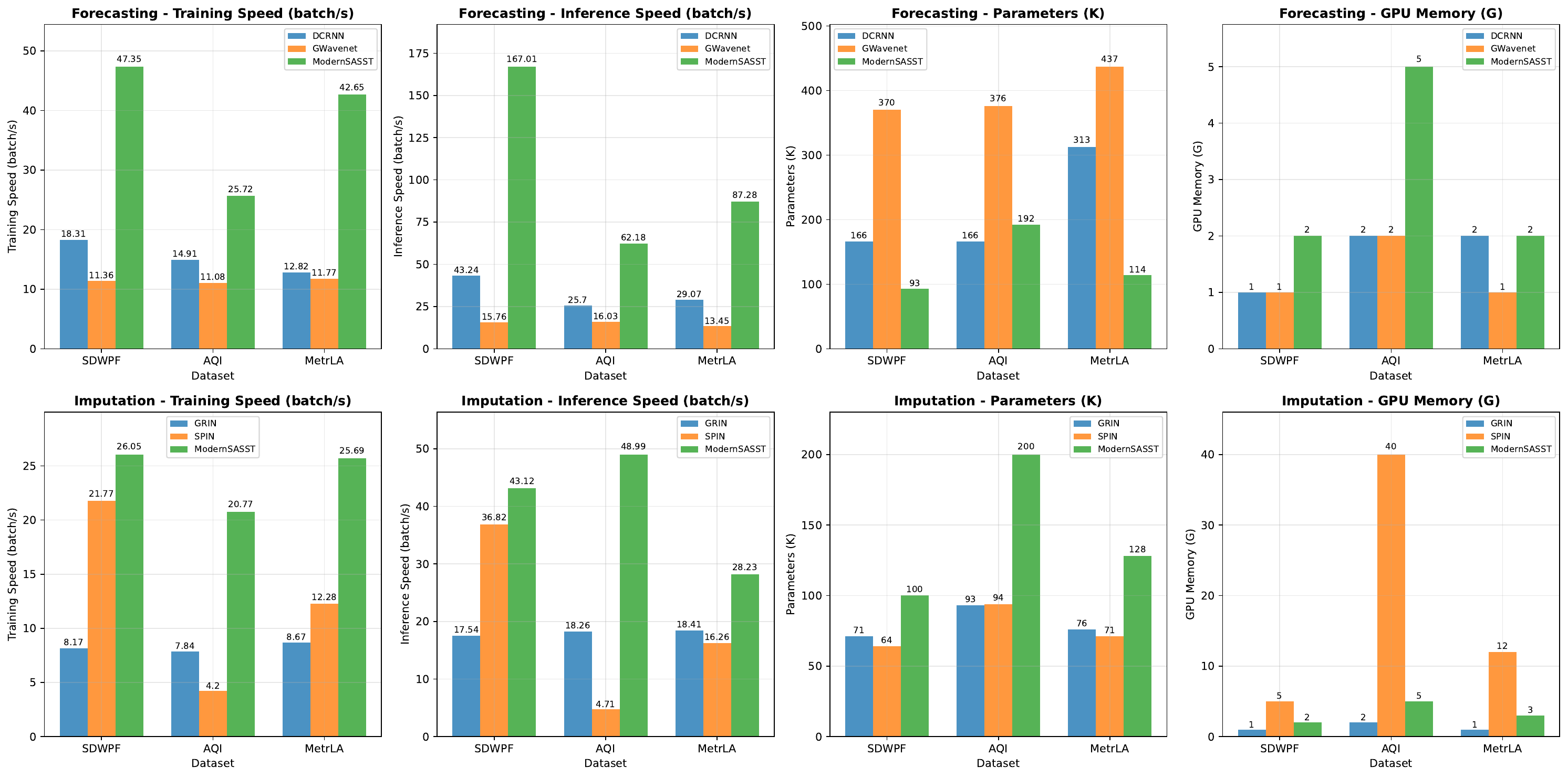}
    \caption{Computational Time}
    \label{fig:computational_time}
\end{figure*}

\begin{figure}[t]
    \centering
    \includegraphics[width=\columnwidth]{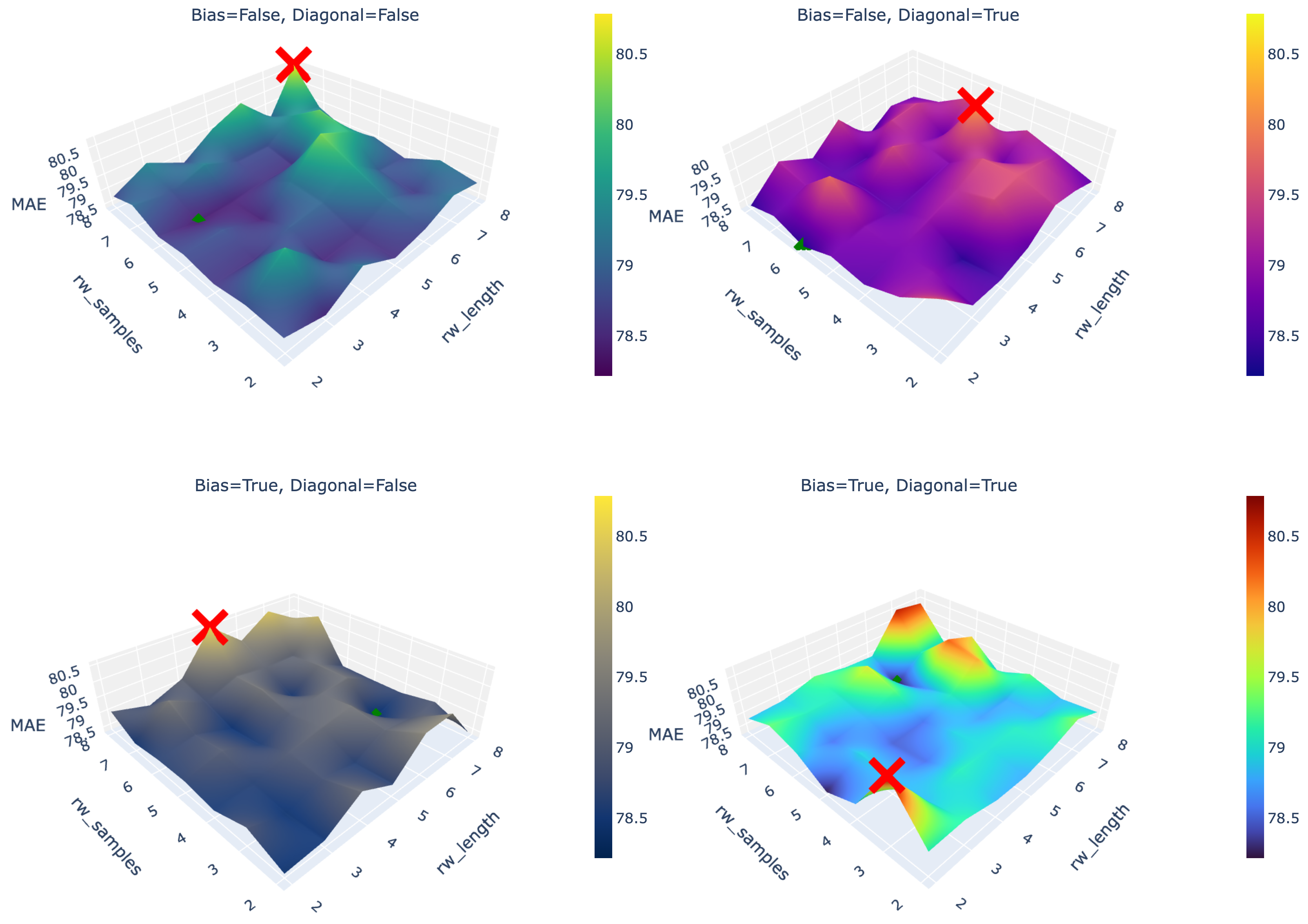}
    \caption{Sensitivity Analysis of SDWPF}
    \label{fig:sensitive_analysis}
\end{figure}

\begin{table}[htbp]
\begin{adjustbox}{width=\columnwidth,center}
\centering
\begin{tabular}{l|cc|cc}
\toprule
& \multicolumn{2}{c|}{SDWPF Forecasting} & \multicolumn{2}{c}{SDWPF Imputation} \\
\cmidrule{2-5}
& MAE & RMSE & MAE & MRE (\%) \\
\midrule
w/o anonymous & 82.36 $\pm$ 0.83 & 151.59 $\pm$ 1.33 & 33.35 $\pm$ 1.54 & 14.22 $\pm$ 0.22 \\
w/o adp-adj \& gated & 79.38 $\pm$ 0.76 & 148.57 $\pm$ 0.52 & 46.54 $\pm$ 2.79 & 19.86 $\pm$ 1.19 \\
w/o Graph-wise LayerNorm & 89.21 $\pm$ 0.31 & 166.34 $\pm$ 0.76 & 30.77 $\pm$ 1.26 & 13.13 $\pm$ 0.54 \\
\midrule
\textbf{Ours} & 79.63 $\pm$ 0.78 & 148.97 $\pm$ 1.11 & 30.49 $\pm$ 1.74 & 13.01 $\pm$ 0.75\\
\bottomrule
\end{tabular}
\end{adjustbox}
\caption{Ablation study on SDWPF dataset}
\label{tab:ablation_results}
\end{table}

\subsection{Sensitive Analysis}
We conduct sensitivity analysis on two key hyperparameters for random walk sampling on the SDWPF dataset. The diagonal parameter controls whether random walks can perform same-order transitions, where $\mathcal{\hat{A}}_1$ includes diagonal terms enabling same-order simplices transitions and $\mathcal{\hat{A}}_2$ constrains transitions to different order simplices only. The bias parameter introduces order-aware transition probabilities to prevent walkers from being trapped in dominant structures. Our experiments evaluate four parameter combinations across varying random walk lengths and sample sizes. The results demonstrate remarkable stability with coefficients of variation ranging from 0.53\% to 0.68\%. The current configuration (diagonal=True, bias=False) exhibits the lowest variation at 0.53\%, with all configurations maintaining mean MAE values within a narrow band around 79.0, indicating minimal sensitivity to these hyperparameter choices. Given the inherent stochasticity of random walk-based methods and their additional challenges in spatiotemporal prediction tasks, we recommend prioritizing robust parameter regions while employing light fine-tuning with fixed learning rates to mitigate sampling-induced variability and maintain the superior performance ceiling demonstrated by optimal configurations.

\subsection{Computational time}
The bar charts present comparative performance metrics across four dimensions: training speed, inference speed, number of parameters, and GPU memory usage. As shown in Figure~\ref{fig:computational_time}, ModernSASST consistently demonstrates superior computational performance across both forecasting and imputation tasks by replacing expensive recurrent mechanisms with parallelizable architectures. The results indicate that ModernSASST successfully balances computational efficiency with resource utilization, making it particularly suitable for real-time spatiotemporal applications.

\section{Discussion and conclusion}
We present ModernSASST, the first framework to leverage simplicial complex structures for spatiotemporal modeling. Our approach uniquely combines spatiotemporal random walks on high-dimensional simplicial complexes with parallelizable Temporal Convolutional Networks, capturing higher-order topological dependencies that go beyond traditional pairwise relationships. By replacing computationally expensive Simplicial Neural Networks with efficient random walk sampling, ModernSASST maintains topological expressiveness while achieving significant computational efficiency gains. This work establishes simplicial complexes as a promising direction for spatiotemporal modeling and opens new research avenues in topological deep learning.

\bigskip

\bibliography{aaai2026}

@inproceedings{defferrard2016convolutional,
  title={Convolutional neural networks on graphs with fast localized spectral filtering},
  author={Defferrard, Micha{\"e}l and Bresson, Xavier and Vandergheynst, Pierre},
  booktitle={Advances in Neural Information Processing Systems},
  volume={29},
  year={2016}
}

@article{kipf2016semi,
  title={Semi-supervised classification with graph convolutional networks},
  author={Kipf, Thomas N and Welling, Max},
  journal={arXiv preprint arXiv:1609.02907},
  year={2016}
}

@inproceedings{veličković2018graph,
  title={Graph Attention Networks},
  author={Petar Veličković and Guillem Cucurull and Arantxa Casanova and Adriana Romero and Pietro Liò and Yoshua Bengio},
  booktitle={International Conference on Learning Representations},
  year={2018}
}

@inproceedings{li2018diffusion,
  title={Diffusion Convolutional Recurrent Neural Network: Data-Driven Traffic Forecasting},
  author={Yaguang Li and Rose Yu and Cyrus Shahabi and Yan Liu},
  booktitle={International Conference on Learning Representations},
  year={2018}
}

@article{wu2019graph,
  title={Graph wavenet for deep spatial-temporal graph modeling},
  author={Wu, Zonghan and Pan, Shirui and Long, Guodong and Jiang, Jing and Zhang, Chengqi},
  journal={arXiv preprint arXiv:1906.00121},
  year={2019}
}

@article{gao2021equivalence,
  title={On the equivalence between temporal and static graph representations for observational predictions},
  author={Gao, Jianfei and Ribeiro, Bruno},
  journal={arXiv preprint arXiv:2103.07016},
  year={2021}
}

@article{battiston2020,
  title={Networks beyond pairwise interactions: Structure and dynamics},
  author={Battiston, Federico and Cencetti, Giulia and Iacopini, Iacopo and Latora, Vito and Lucas, Maxime and Patania, Alice and Young, Jean-Gabriel and Petri, Giovanni},
  journal={Physics Reports},
  volume={874},
  pages={1--92},
  year={2020}
}

@article{bick2023,
  title={What are higher-order networks?},
  author={Bick, Christian and Gross, Elizabeth and Harrington, Heather A and Schaub, Michael T},
  journal={SIAM Review},
  volume={65},
  number={3},
  pages={686--731},
  year={2023}
}

@misc{hajij2023topological,
  title={Topological Deep Learning: Going Beyond Graph Data},
  author={Mustafa Hajij and others},
  year={2023},
  eprint={2206.00606},
  archivePrefix={arXiv}
}

@misc{papillon2023architectures,
  title={Architectures of Topological Deep Learning: A Survey on Topological Neural Networks},
  author={Mathilde Papillon and Sophia Sanborn and Mustafa Hajij and Nina Miolane},
  year={2023},
  eprint={2304.10031},
  archivePrefix={arXiv}
}

@article{schaub2020random,
  title={Random walks on simplicial complexes and the normalized Hodge 1-Laplacian},
  author={Schaub, Michael T and Benson, Austin R and Horn, Paul and Lippner, Gabor and Jadbabaie, Ali},
  journal={SIAM Review},
  volume={62},
  number={2},
  pages={353--391},
  year={2020}
}

@inproceedings{bodnar2021weisfeiler,
  title={Weisfeiler and lehman go topological: Message passing simplicial networks},
  author={Bodnar, Cristian and Frasca, Fabrizio and Wang, Yuguang and Otter, Nina and Montufar, Guido F and Lio, Pietro and Bronstein, Michael},
  booktitle={International Conference on Machine Learning},
  pages={1026--1037},
  year={2021}
}

@inproceedings{bunch2020simplicial,
  title={Simplicial 2-Complex Convolutional Neural Networks},
  author={Eric Bunch and Qian You and Glenn Fung and Vikas Singh},
  booktitle={TDA {\&} Beyond Workshop, NeurIPS},
  year={2020}
}

@inproceedings{gurugubelli2024sann,
  title={Sa{NN}: Simple Yet Powerful Simplicial-aware Neural Networks},
  author={Gurugubelli, Sravanthi and Chepuri, Sundeep Prabhakar},
  booktitle={International Conference on Learning Representations},
  year={2024}
}

@inproceedings{luo2024moderntcn,
  title={ModernTCN: A modern pure convolution structure for general time series analysis},
  author={Luo, Donghao and Wang, Xue},
  booktitle={International Conference on Learning Representations},
  year={2024}
}

@article{bai2018empirical,
  title={An empirical evaluation of generic convolutional and recurrent networks for sequence modeling},
  author={Bai, Shaojie and Kolter, J Zico and Koltun, Vladlen},
  journal={arXiv preprint arXiv:1803.01271},
  year={2018}
}

@inproceedings{perozzi2014deepwalk,
  title={DeepWalk: Online learning of social representations},
  author={Perozzi, Bryan and Al-Rfou, Rami and Skiena, Steven},
  booktitle={Proceedings of the 20th ACM SIGKDD},
  pages={701--710},
  year={2014}
}

@inproceedings{grover2016node2vec,
  title={node2vec: Scalable feature learning for networks},
  author={Grover, Aditya and Leskovec, Jure},
  booktitle={Proceedings of the 22nd ACM SIGKDD},
  pages={855--864},
  year={2016}
}

@inproceedings{tang2015line,
  title={Line: Large-scale information network embedding},
  author={Tang, Jian and others},
  booktitle={Proceedings of the 24th WWW},
  pages={1067--1077},
  year={2015}
}

@article{billings2019simplex2vec,
  title={Simplex2vec embeddings for community detection in simplicial complexes},
  author={Billings, Jacob Charles Wright and others},
  journal={arXiv preprint arXiv:1906.09068},
  year={2019}
}

@article{hacker2020k,
  title={k-simplex2vec: a simplicial extension of node2vec},
  author={Hacker, Celia},
  journal={arXiv preprint arXiv:2010.05636},
  year={2020}
}

@article{ebli2020simplicial,
  title={Simplicial neural networks},
  author={Ebli, Stefania and Defferrard, Micha{\"e}l and Spreemann, Gard},
  journal={arXiv preprint arXiv:2010.03633},
  year={2020}
}

@article{wang2025non,
  title={Non-convolutional graph neural networks},
  author={Wang, Yuanqing and Cho, Kyunghyun},
  journal={Advances in Neural Information Processing Systems},
  volume={37},
  year={2025}
}

@inproceedings{chung2014empirical,
  title={Empirical evaluation of gated recurrent neural networks on sequence modeling},
  author={Chung, Junyoung and Gulcehre, Caglar and Cho, Kyunghyun and Bengio, Yoshua},
  booktitle={NIPS Deep Learning Workshop},
  year={2014}
}

@article{wang2019deep,
  title={Deep graph library: A graph-centric, highly-performant package for graph neural networks},
  author={Wang, Minjie and others},
  journal={arXiv preprint arXiv:1909.01315},
  year={2019}
}

@article{micali2016reconstructing,
  title={Reconstructing markov processes from independent and anonymous experiments},
  author={Micali, Silvio and Zhu, Zeyuan Allen},
  journal={Discrete Applied Mathematics},
  volume={200},
  pages={108--122},
  year={2016}
}

@inproceedings{Yuqietal-2023-PatchTST,
  title={A Time Series is Worth 64 Words: Long-term Forecasting with Transformers},
  author={Nie, Yuqi and Nguyen, Nam H and Sinthong, Phanwadee and Kalagnanam, Jayant},
  booktitle={International Conference on Learning Representations},
  year={2023}
}

@article{ba2016layernormalization,
  title={Layer Normalization},
  author={Ba, Jimmy Lei and Kiros, Jamie Ryan and Hinton, Geoffrey E},
  journal={arXiv preprint arXiv:1607.06450},
  year={2016}
}

@article{zhou2024sdwpf,
  title={SDWPF: a dataset for spatial dynamic wind power forecasting over a large turbine array},
  author={Zhou, Jingbo and others},
  journal={Scientific Data},
  volume={11},
  number={1},
  pages={649},
  year={2024}
}

@book{hamilton2020time,
  title={Time series analysis},
  author={Hamilton, James D},
  year={2020},
  publisher={Princeton University Press}
}

@article{sutskever2014sequence,
  title={Sequence to sequence learning with neural networks},
  author={Sutskever, Ilya and Vinyals, Oriol and Le, Quoc V},
  journal={Advances in Neural Information Processing Systems},
  volume={27},
  year={2014}
}

@article{cini2023scalable, 
  title={Scalable Spatiotemporal Graph Neural Networks}, 
  journal={Proceedings of the AAAI Conference on Artificial Intelligence},
  author={Cini, Andrea and Marisca, Ivan and Bianchi, Filippo Maria and Alippi, Cesare}, 
  volume={37},
  number={6},
  pages={7218-7226},
  year={2023} 
}

@article{cao2018brits,
  title={BRITS: Bidirectional Recurrent Imputation for Time Series},
  author={Cao, Wei and others},
  journal={Advances in Neural Information Processing Systems},
  volume={31},
  year={2018}
}

@inproceedings{cini2022filling,
  title={Filling the G\_ap\_s: Multivariate Time Series Imputation by Graph Neural Networks},
  author={Cini, Andrea and Marisca, Ivan and Alippi, Cesare},
  booktitle={International Conference on Learning Representations},
  year={2022}
}

@article{marisca2022learning,
  title={Learning to reconstruct missing data from spatiotemporal graphs with sparse observations},
  author={Marisca, Ivan and Cini, Andrea and Alippi, Cesare},
  journal={Advances in Neural Information Processing Systems},
  volume={35},
  pages={32069--32082},
  year={2022}
}

@inproceedings{akiba2019optuna,
  title={Optuna: A next-generation hyperparameter optimization framework},
  author={Akiba, Takuya and others},
  booktitle={Proceedings of the 25th ACM SIGKDD},
  year={2019}
}

@inproceedings{dong2017metapath2vec,
  title={metapath2vec: Scalable representation learning for heterogeneous networks},
  author={Dong, Yuxiao and Chawla, Nitesh V and Swami, Ananthram},
  booktitle={Proceedings of the 23rd ACM SIGKDD international conference on knowledge discovery and data mining},
  pages={135--144},
  year={2017}
}

@inproceedings{he2019hetespaceywalk,
  title={Hetespaceywalk: A heterogeneous spacey random walk for heterogeneous information network embedding},
  author={He, Yu and Song, Yangqiu and Li, Jianxin and Ji, Cheng and Peng, Jian and Peng, Hao},
  booktitle={Proceedings of the 28th ACM international conference on information and knowledge management},
  pages={639--648},
  year={2019}
}
\bibliographystyle{plainnat} % 或者使用 abbrvnat

\end{document}